\documentclass[letterpaper, 10 pt, conference]{ieeeconf}  % Comment this line out if you need a4paper
\IEEEoverridecommandlockouts % This command is only needed if you want to use the \thanks command
\usepackage{epsfig} % for postscript graphics files 
\DeclareMathAlphabet{\mathcal}{OMS}{cmsy}{m}{n}

\usepackage[T1]{fontenc} 
\usepackage[utf8]{inputenc}
\usepackage[english]{babel}
\DeclareUnicodeCharacter{064D}{}

\usepackage{graphicx}   
\usepackage{mathptmx}   

\usepackage{makecell}
\usepackage{tabularx}
\usepackage{adjustbox}
\usepackage{multirow}
\usepackage{array}
\newcolumntype{P}[1]{>{\centering\arraybackslash}p{#1}}
\usepackage{booktabs} 

\usepackage{amsmath} % assumes amsmath package installed
\usepackage{amssymb}  % assumes amsmath package installed
\usepackage{bm}

\usepackage[version-1-compatibility]{siunitx}

\let\labelindent\relax
\usepackage{enumitem}
\usepackage[dvipsnames]{xcolor}
\usepackage{soul} % for drafting

\usepackage{cite}
\usepackage[hyphens]{url}
\usepackage[hidelinks]{hyperref}
\hypersetup{breaklinks=true}

\usepackage{comment}  
\usepackage{xstring}
\usepackage{forloop}

\usepackage{algorithm}
\usepackage{algpseudocode}

\makeatletter
\newcommand*\bigcdot{\mathpalette\bigcdot@{1}}
\newcommand*\bigcdot@[2]{\mathbin{\vcenter{\hbox{\scalebox{#2}{$\m@th#1\bullet$}}}}}
\makeatother

\usepackage{caption}
\begin{document}

\newcommand\blfootnote[1]{%
  \begingroup
  \renewcommand\thefootnote{}\footnote{#1}%
  \addtocounter{footnote}{-1}%
  \endgroup
}

\title{\LARGE \bf
 DSP-SLAM++: A Unified Framework for Multi-Class, High-Fidelity \\  Object SLAM in the Wild
%Option2:: MC-DSP-SLAM: Multi-Class Object SLAM with Deep Shape Priors for Asynchronous Monocular Fisheye-LiDAR Systems
}

%%% Authors
\author{Ahmad Kourani$^{1}$, Ghina Daoud$^{1}$, Daniel Asmar$^{2}$, and Imad Elhajj$^{3}$% <-this % stops a space
\thanks{*This work was supported by the DIDMOS-XR project.}% <-this % stops a space
\thanks{$^{1}$Ahmad Kourani and Ghina Daoud are with the Vision and Robotics Lab (VRL), 
        American University of Beirut (AUB), Beirut, Lebanon 
        {\tt\small \{ahk42, gad08\}@mail.aub.edu}}%
\thanks{$^{2}$Daniel Asmar is with the Department of Mechanical Engineering, MSFEA, 
        American University of Beirut (AUB), Beirut, Lebanon 
        {\tt\small da20@aub.edu.lb}}%
\thanks{$^{3}$Imad Elhajj is with the Department of Electrical and Computer Engineering, MSFEA, 
        American University of Beirut (AUB), Beirut, Lebanon 
        {\tt\small ie05@aub.edu.lb}}%
}

\maketitle
\thispagestyle{empty}
\pagestyle{empty}

% -----------------------------------------------------------------------------------------
\begin{abstract}
Existing object-aware SLAM systems force a trade-off between real-time performance, multi-class support, and the generation of high-fidelity, semantically coherent object models. 
To address this trade-off, we present DSP-SLAM++, which extends the DSP-SLAM framework with an asynchronous mapping pipeline for real-time performance and dedicated sensor fusion adaptations for a monocular fisheye-LiDAR suite.
Experiments demonstrate that our system generates fine-grained, geometrically-complete shapes for multiple object classes while eliminating severe mapping thread bottlenecks by reducing maximum object processing latency by up to 70\% compared to the state-of-the-art baseline, enabling robust, real-time performance on a challenging 25 Hz multi-class datasets.
This work makes high-fidelity, multi-class object SLAM more practical for real-world applications like autonomous driving and robotic manipulation by enabling its use on platforms with common fisheye-LiDAR sensor setups.
The open-source code is available at: [github.com/AUBVRL/DSP-SLAMpp].
\end{abstract}

% \blfootnote{
% $^{1}$Ahmad Kourani is with the Mechanical Engineering Department, Vision and Robotics Lab, American University of Beirut, Beirut, Lebanon, {\tt\small ahk42@mail.aub.edu} \\

%\indent $^{2}$Daniel Asmar is with the Mechanical Engineering Department, Vision and Robotics Lab, American University of Beirut, Beirut, Lebanon,
%     {\tt\small da20@aub.edu.lb}}%

\begin{keywords}
Object-SLAM, Deep Shape Priors, Multi-Class Reconstruction, Asynchronous LiDAR-Camera Fusion, Fisheye SLAM.
\end{keywords}

%%%%%%%%%%%%%%%%%%%%%%%%%%%%%%%%%%%%%%%%%%%%%%%%%%%%%%%%%%%%%%%%%%%%%%
%%%%%%%%%%%%%%%%%%%%%%%%%%%%%%%%%%%%%%%%%%%%%%%%%%%%%%%%%%%%%%%%%%%%%%
\section{Introduction} 
\label{sec:introduction}
%
%
%
%
%
% To navigate safely in complex road environment, the perception systems of autonomous vehicles must achieve a deep, object-level understanding of their surroundings. To this end, the field of Simultaneous Localization and Mapping (SLAM) has progressed beyond simple geometric maps to the frontier of object-aware SLAM, which models individual objects as persistent entities. A key challenge in most current methods is that they represent these objects using simplified primitives like cubic boxes \cite{cubeslam,eaoslam} or quadrics \cite{nicholson2019qslam,tian2022quadric}. While computationally efficient, these low-fidelity representations are insufficient for capturing the precise geometry of real-world objects. As a result, this limits the ability to accurately forecast an object's future motion and plan safe, nuanced maneuvers in close-proximity interactions, creating a clear need for richer representations that can model complex object shapes with high fidelity.
%-------------------------------------------------
% Real-world autonomous driving and robotic applications must handle object categories of varying size, geometry, appearance and semantic importance \cite{caesar2020nuscenes}, including pedestrians, cars, buses, and motorcycles, where reliable multi-class reconstruction and tracking is critical for reliability and safety. 
For autonomous vehicles to navigate safely in complex on-road environments, their perception systems must achieve a deep, object-level understanding of their surroundings. To this end, the field of Simultaneous Localization and Mapping (SLAM) has progressed from simple geometric maps to the frontier of object-aware SLAM, which models individual objects as persistent, semantic entities. A key challenge in current methods lies in their reliance on simplified primitives, such as cubic boxes \cite{cubeslam,eaoslam} or quadrics \cite{nicholson2019qslam,tian2022quadric}, to represent objects. While computationally efficient, these low-fidelity representations fail to capture the precise geometry of real-world items, which hinders the ability to accurately forecast object motion and plan safe, nuanced maneuvers during close-proximity interactions. Consequently, a clear need exists for richer map representations capable of modeling complex object shapes with high fidelity.
%
% Safe navigation for autonomous vehicles requires the reliable reconstruction and tracking of diverse object categories—such as pedestrians, cars, and buses—that vary in size, geometry, and semantic importance. Achieving this deep, multi-class object understanding is a primary goal of modern object-aware SLAM, which progresses beyond simple geometric maps to model individual objects as persistent, semantic entities. A key challenge in current methods lies in their reliance on simplified primitives—such as cubic boxes \cite{cubeslam,eaoslam} or quadrics \cite{nicholson2019qslam,tian2022quadric}—to represent objects. While computationally efficient, these low-fidelity representations fail to capture the precise geometry of real-world items, hindering the ability to accurately forecast object motion and plan safe, nuanced maneuvers during close-proximity interactions. Consequently, a clear need exists for richer map representations capable of modeling complex object shapes with high fidelity.
% -------------
\par
% For autonomous vehicles to navigate safely in complex on-road environments, their perception systems must achieve a deep, object-level understanding of their surroundings. While the field of object-aware SLAM has progressed significantly, current approaches force a trade-off between three critical capabilities. Some systems can generate high-fidelity shapes but are computationally expensive and often limited to single object classes. Others achieve real-time, multi-class performance but rely on coarse geometric primitives that fail to capture precise object geometry. A third category offers visually impressive reconstructions but lacks a coherent, instance-level semantic understanding of objects. This forces a compromise between geometric accuracy, speed, and semantic coherence, which hinders downstream tasks like motion forecasting and safe interaction planning. Consequently, a significant gap remains for a unified system that can simultaneously deliver high-fidelity, multi-class object reconstruction in real-time.
%-------------------- begin figure-------------------

\begin{figure}[t]
    \centering
    \includegraphics[width=\linewidth]{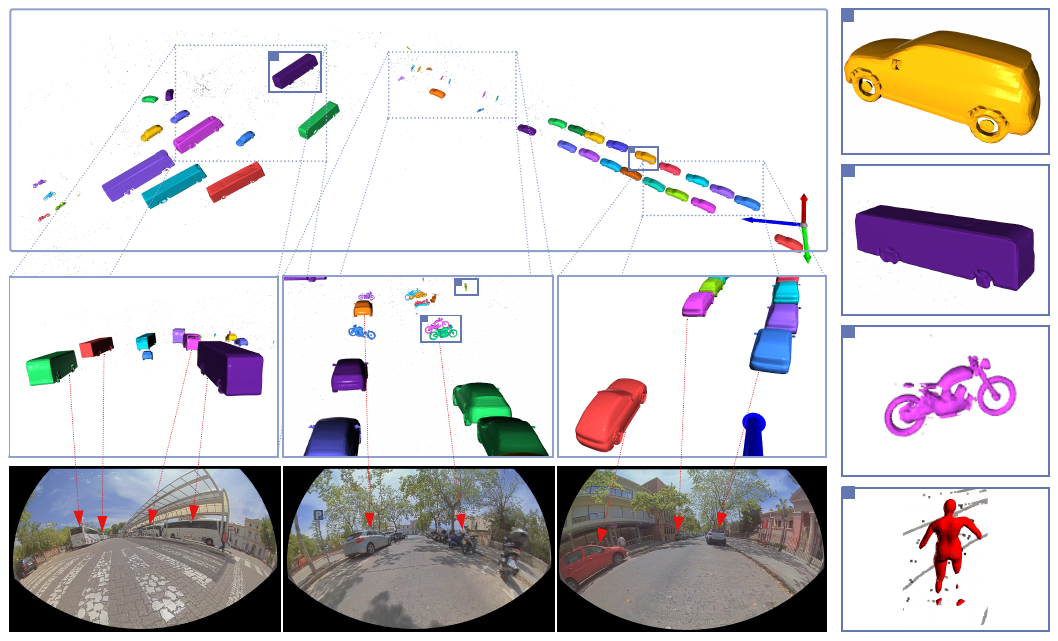}
    \caption{DSP-SLAM++ generates consistent object-aware maps rich with multi-class objects of detailed shapes in real time. Reconstructed map from our in-house dataset.}
    \label{fig:custom_dataset_map}
\end{figure}
%-------------------- end figure-------------------
\par
%-------------------- begin figure-------------------
%%% 3.34in is the maximum width you can have for a figure
\begin{figure*} [h]
\centerline{\includegraphics[width=7 in]{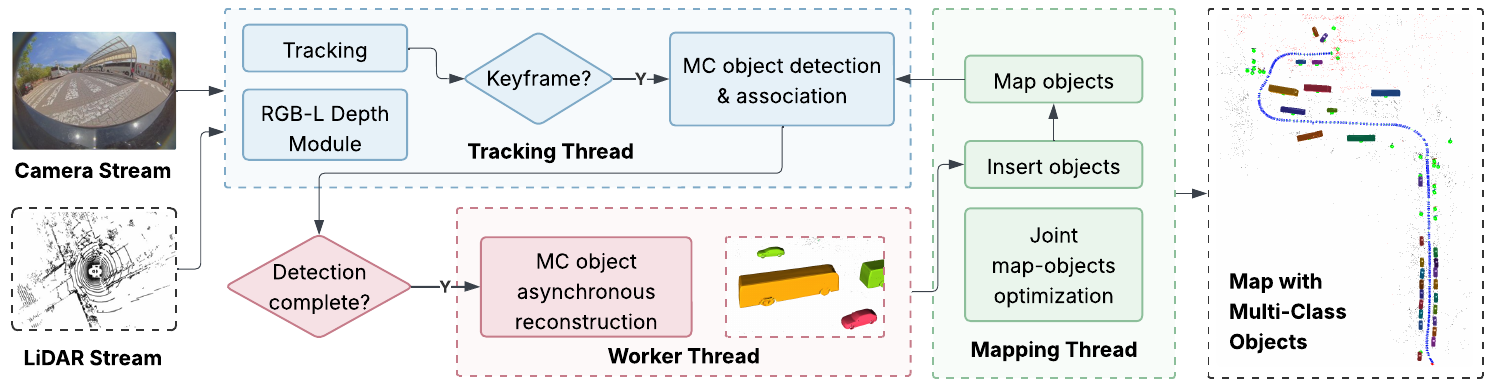}}
\caption{DSP-SLAM++ system overview. MC stands for Multi-Class.}
\label{fig:system_overview}
\end{figure*}
%-------------------- end figure-------------------
%
% \textcolor{blue}{Priming: Detailed Shapes, and DSP-SLAM Limitations}
% \par
% More advanced systems like DSP-SLAM \cite{wang2021dsp} leverage category-specific deep shape embeddings known as DeepSDF \cite{park2019deepsdf} as priors to reconstruct detailed and complete 3D mesh even from partial observations, while jointly optimizing camera poses, object location, and sparse landmarks in the map. Despite its effectiveness, DSP-SLAM is restricted, in both its single object and multiple object implementations, to a single object class, which limits its applicability in diverse outdoor environments. Additionally, it does not generalize and integrate well to real life setups, as it relies on stereo camera for depth perception even in the presence of LiDAR senor, supports only pinhole type camera lens, assumes synchronized Camera and LiDAR streams, and suffer from a computational speed bottleneck when the number of objects increase.
\par
To address the need for high-fidelity shapes, systems like DSP-SLAM \cite{wang2021dsp} leverage category-specific deep shape priors to reconstruct detailed 3D meshes from partial observations. However, its heavy computational cost and restrictive implementation limited its applicability to a single class of objects and narrow sensor suites. A more recent paradigm attempts to solve the speed and visual quality problem using explicit representations like 3D Gaussian Splatting, but lacks a coherent, instance-level understanding of objects \cite{zhu2024semgauss}. This leaves a critical gap for a unified system that combines geometric detail with real-time speed and semantic coherence required for robust, object-level autonomy.
\par
% ----------------------------------------------
% --------------------------------------------------------
% --------------------------------------------------------
% --------------------------------------------------------
%
% \textcolor{blue}{Proposed system}
% \par
%
To address the challenges of object-aware SLAM, we present DSP-SLAM++ (Fig.~\ref{fig:custom_dataset_map}), a unified framework designed for the complexities of real-world automotive setups. Our system builds upon the implicit shape representations of DSP-SLAM but redesigns its architecture for real-time, multi-class performance through a novel asynchronous pipeline and a dedicated sensor fusion module. This approach allows our system to generate high-fidelity, multi-class object maps while running at 25 Hz without performance or stability degradation compared to the state-of-the-art baseline.
\par
The main contributions of this work are:
\begin{enumerate}
    \item The development of a real-time SLAM system that reconstructs multiple classes of objects with high geometric fidelity, providing richer, more semantically diverse maps for downstream robotics or tasks like motion forecasting and planning.
    \item A novel asynchronous processing pipeline that resolves the computational bottlenecks of prior implicit methods, enabling robust real-time performance regardless of the number of objects being modeled.
    \item A complete sensor fusion and compensation pipeline that enables the use of a challenging but practical monocular fisheye and LiDAR sensor suite, making high-fidelity object SLAM more accessible to common automotive platforms by leveraging stock camera installations found in high-end vehicles.
    \item The release of our open-source implementation to the community to foster further research and development.
\end{enumerate}
\par
\section{Related Work}
\label{sec:realted_work}
\subsection{Multi-Class Object Reconstruction}

Object representation in SLAM traditionally forces a trade-off between geometric fidelity and computational efficiency. Approaches utilizing learned implicit shape priors, such as DSP-SLAM \cite{wang2021dsp}, achieve detailed 3D mesh reconstructions but suffer from high computational costs and single-class restrictions. Conversely, modeling objects with lightweight geometric primitives—such as cuboids \cite{cubeslam,eaoslam} or ellipsoids \cite{tian2022quadric,pan2025mcoo}—ensures real-time, multi-class scalability but sacrifices the precise shape details required for complex tasks. 
\par
A recent paradigm shift has emerged with 3D Gaussian Splatting (3DGS) \cite{wu2024_3DGS}, which models the environment as a collection of explicit primitives, enabling high-fidelity, photo-realistic rendering in real-time. While advanced systems can augment individual Gaussians with semantic labels \cite{zhu2024semgauss}, they fundamentally lack a coherent, instance-level understanding of objects, representing them as a collection of points rather than as distinct geometric entities.

\subsection{Sensor Fusion and Compensation for Automotive SLAM}

Transitioning advanced SLAM algorithms to real-world automotive platforms introduces challenges related to practical sensor configurations. The narrow Field-of-View (FoV) of traditional pinhole cameras creates unacceptable blind spots, leading to a trend of adopting wide-angle fisheye cameras in modern datasets and vehicles~\cite{yogamani2019woodscape, liao2022kitti, yang2024mcov}. In response, state-of-the-art geometric SLAM frameworks like ORB-SLAM3 have incorporated unified camera models to handle the high distortion inherent in these lenses~\cite{campos2021orb}, highlighting a key limitation in prior object-aware systems that lack native fisheye support.

Another significant challenge is the inherent scale ambiguity of monocular visual SLAM. While stereo cameras can resolve this issue, they add cost, complexity, and redundancy to systems already equipped with a LiDAR sensor~\cite{badue2021carsSurvey}. Consequently, LiDAR-assisted visual SLAM has been explored to provide metric scale. Some methods fuse LiDAR depth with feature tracks and semantic priors~\cite{limo2018, led2024}, though often at a high computational cost on GPUs. Others are tailored to specific hardware setups~\cite{camvox2020}. In contrast, lightweight, CPU-based alternatives like RGB-L~\cite{rgbl2022} directly leverage LiDAR point clouds to recover scale and improve accuracy, offering a practical approach for real-world deployment.

% Finally, the asynchronous capture rates of different sensors on a moving vehicle cause spatial misalignment between their respective frames. A LiDAR scan captured at one instant will appear shifted it time relative to the nearest camera frame ~\cite{yuan2022licas3}. The standard approach to correct this issue is to apply ego-motion compensation, which uses the vehicle's trajectory and sensor timestamps to compute the transformation that occurs between capture times, and then align the frames to a common reference point~\cite{rehder2016general}.

% -----------------------------------------
% -----------------------------------------
% -----------------------------------------
In summary, the literature reveals a two-fold challenge for practical deployment of object-aware SLAM in the wild. The field of object representation is fragmented by trade-offs between geometric accuracy, performance, and semantic coherence. Furthermore, real-world automotive platforms introduce significant sensor fusion and compensation complexities. A robust solution must therefore unify these competing representational goals while mastering the challenges of asynchronous, wide-angle sensors. Motivated by these challenges, our work proposes a system that integrates the expressive power of deep implicit models into a real-time, multi-class architecture.

\section{System Overview} 
\label{sec:System_Overview}  
The overall architecture of our proposed DSP-SLAM++ is illustrated in Figure~\ref{fig:system_overview}. It is a unified framework that takes monocular fisheye images and LiDAR scans as input, and produces a globally consistent trajectory, alongside a map of high-fidelity, multi-class object reconstructions. The system operates through several parallel threads: a front-end for tracking and sensor fusion, an object detection and association module, and a back-end for asynchronous reconstruction and global optimization.

The front-end thread 
% first performs ego-motion compensation to spatiotemporally align the asynchronous sensor streams. It then 
leverages a lightweight RGB-L module to generate a dense depth map from the aligned LiDAR data, providing a metric scale to the monocular SLAM pipeline. These components, detailed in Sec.~\ref{sec:lidar_cam_sync}, allow the ORB-SLAM3 backbone to estimate robust camera poses using its native fisheye camera support. In parallel, an object detection module identifies and associates objects from the sensor data (Sec.~\ref{sec:Multi_Class}). Data for newly observed objects is dispatched to our asynchronous reconstruction back-end, which uses a thread pool and category-specific DeepSDF networks to generate detailed 3D meshes without blocking the main tracking thread (Sec.~\ref{sec:asynch_recons}). Finally, the estimated camera and objects poses, and sparse map points are jointly refined to maintain global map consistency.
\section{Multi-Class Detection \& Reconstruction} 
\label{sec:Multi_Class}  
% start writing here
% This section details the workflow for creating a rich, multi-class object map, illustrated in Fig.~\ref{fig:multi_class_system}. The primary goal is to produce semantically meaningful object landmarks that are jointly optimized with the camera trajectory and map points. This is accomplished in three sequential stages: object detection and initialization from raw sensor data, multi-class data association against existing map entities, and finally, class-specific shape reconstruction using a set of dedicated DeepSDF networks, one for each object class.
%
Our multi-class object reconstruction pipeline, illustrated in Algorithm~\ref{alg:multi_class_pipeline}, produces semantically meaningful landmarks that are jointly optimized with the camera trajectory and map points. The workflow consists of three sequential stages: object detection and initialization, multi-class data association, and class-specific shape reconstruction. 

\par
% \textcolor{red}{Add an algorithm that explains the MC framework.}
% ---------------------------------------------------------
% ---------------------------------------------------------
% ---------------------------------------------------------
% start writing here
% \textbf{A. Object Detection} \\
\subsection{Object Candidate Generation}

The first stage of our pipeline generates object candidates at each keyframe in a similar manner as the baseline DSP-SLAM. We perform 2D object detection using YOLO \cite{yolo11_ultralytics}, where each detected instance $d_{2D,j}$ in $\mathcal{D}_{2D}$ with $j \in \{1,...,n_{2D}\}$ provides a segmentation mask $\mathcal{M}_{2D,j}$ and a class label $c_{2D,j}$ (Line 2). In parallel, 3D detections are obtained via PointPillars~\cite{lang2019pointpillars}, with each instance $d_{3D,j}$ of $\mathcal{D}_{3D}$ with $i \in \{1,...,n_{3D}\}$ providing a class-labeled 3D bounding box $b_{3D,i}$. We filter the raw 2D and 3D detections using class-specific confidence thresholds ($\tau_{c,2D}$, $\tau_{c,3D}$). 
\par
%
% By jointly leveraging 2D and 3D detectors and calibrating them per category, this stage provides consistent, class-aware observations that form the foundation for the multi-class SLAM pipeline.
% \begin{figure}[t]
%   \centering
%   \includegraphics[width=1.0\linewidth]{figures/multi_class_system.png}
%   \caption{Temporary overview of the multi-class support in the DSP-SLAM framework \textcolor{blue}{(to be updated)}}
%   \label{fig:multi_class_system}
% \end{figure}
%
\begin{algorithm}[t]
\caption{Multi-Class Object Reconstruction Pipeline}
\label{alg:multi_class_pipeline}
\begin{algorithmic}[1]
    \Procedure{ProcessKeyframeObjects}{image, lidar\_scan}
        \Statex \Comment{Stage 1: Per-Frame Candidate Generation}
        \State $(\mathcal{D}_{2D}, \mathcal{D}_{3D}) \gets (\text{Det}_{2D}(\text{image}), \text{Det}_{3D}(\text{velo}))$
        \State $\mathcal{D}_{\text{valid}} \gets \text{GlobalGreedyAssignment}(\mathcal{D}_{2D}, \mathcal{D}_{3D})$
        
        \Statex \Comment{Stage 2: Class-Aware Map Association}
        \State $\text{unassociated\_candidates} \gets \mathcal{D}_{\text{valid}}$
        \ForAll{object $m \in \text{GlobalMap}$}
            \State $d^* \gets \text{FindBestMatch}(m, \text{unassociated\_candidates})$
            \If{$d^*$ is not \textbf{null}}
                \State AssociateObservation($m, d^*$)
                \State $\text{unassociated\_candidates}$.remove($d^*$)
            \EndIf
        \EndFor

        \Statex \Comment{Initialize new objects from remaining candidates}
        \ForAll{candidate $d \in \text{unassociated\_candidates}$}
            \State $m_{\text{new}} \gets \text{InitializeNewObject}(d)$
            \State TriggerReconstruction($m_{\text{new}}$) 
            \State GlobalMap.add($m_{\text{new}}$)
        \EndFor
    \EndProcedure
\end{algorithmic}
\end{algorithm}

The baseline DSP-SLAM's approach to 2D-3D matching, which exhaustively projects every 3D instance onto every 2D mask, becomes inefficient and prone to wrong associations when extended to multiple object classes. 
To address this limitation, we introduce a robust global greedy bipartite matching strategy designed for the complexities of multi-class scenarios (Line 3). 
Our process first establishes a direct semantic link between detections through a class-consistent projection strategy, and finally, a disambiguation method resolves assignment ambiguities common in complex scenes with nested segmentation masks. This approach ensures a reliable one-to-one matching between 2D and 3D detections.
\par
The assignment process begins by establishing class-consistency between the sets $\mathcal{D}_{2D}$ and $\mathcal{D}_{3D}$. 
Let $\mathcal{P}_{3D,i}$ be the $i^{th}$ 3D detection's point cloud projected onto the image plane; 
we define the association score $|\mathcal{P}_{3D,i} \cap \mathcal{M}_{2D,j}|$ as the ratio of projected points that fall within the mask boundaries to the total number of points. In the case of a fisheye lens, masks are rectified prior to projection to ensure geometric consistency.
We construct a global pairwise score matrix, $S \in \mathbb{Z}^{n_{3D} \times n_{2D}}$, that connects all 3D instances to all 2D masks, while only evaluated for candidate pairs belonging to the same class and invalidated for the rest:
\begin{equation}
    S_{i,j} = 
    \begin{cases} 
    |\mathcal{P}_{3D,i} \cap \mathcal{M}_{2D,j}| & \text{if } c_{3D,i} = c_{2D,j}, \\ 
    -1 & \text{otherwise.} 
    \end{cases}
    \label{eq:score_matrix}
\end{equation}
%
% while eliminating unnecessary computations. 
Assignment ambiguity could happen when a single 3D detection is projected onto multiple tangent masks. To resolve this issue, we employ a global greedy bipartite matching strategy. We iteratively identify the candidate pair $(i^*, j^*)$ that yields the highest inlier score in the entire matrix:
\begin{equation}
    (i^*, j^*) = \underset{i, j}{\arg\max} \, S_{i,j}
    \label{eq:greedy_match}
\end{equation}
This maximum score is then validated against a class-specific point-ratio threshold. 
If a match is successful, the 3D candidate $d_{3D,i^*}$ is assigned to the 2D mask $d_{2D,j^*}$, while simultaneously invalidating the corresponding row $i^*$ and column $j^*$ in $S$:
\begin{equation}
    S_{i^*, j} = -1 \quad \forall j \in [1, n_{2D}], \quad S_{i, j^*} = -1 \quad \forall i \in [1, n_{3D}]
    \label{eq:matrix_invalidation}
\end{equation}
This solution enforces a mutually exclusive, one-to-one mapping of the 3D detections and masks.
The greedy selection process is repeated iteratively until all valid scores $S_{i,j} > 0$ are exhausted. 

Our proposed approach improves the robustness and computational efficiency of the 2D-3D association by resolving assignment ambiguities in a single unified step, yielding a set of validated 2D-3D object pairs for the subsequent map association and reconstruction stages.

% --------------------------------------------------------
% --------------------------------------------------------
% --------------------------------------------------------
% --------------------------------------------------------
\subsection{Class-Aware Map Association}
% --------------------------------
\begin{figure}[t]
  \centering
  \includegraphics[width=1.0\linewidth]{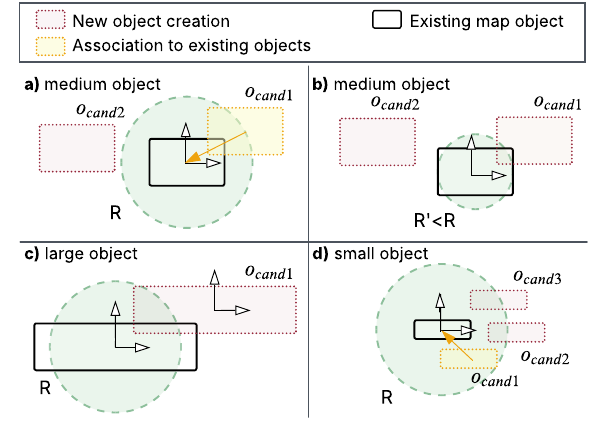}
  \caption{Map objects association for different search radii. a) appropriate search radius (correct association), b) Small search radius (missing association), c) large objects (missing association) d) small objects (wrong association).}
  %\caption{Illustration of the class-conditioned data association mechanism. The figure demonstrates how our class-specific association threshold, $\mathrm{r}_c$, improves tracking reliability over a uniform radius. (a) A new detection (dashed circle) is correctly associated with a map object (solid circle) using an appropriate class-specific radius. (b) A failed association occurs when the search radius is too small for the object's scale. (c) For large objects, our method successfully makes an association by using a larger, geometry-aware radius. (d) In a crowded scene, a small object is protected from false association with a larger neighbor due to its smaller, correctly-scaled radius.}
  \label{fig:map_objects_association}
\end{figure}
% --------------------------------
To track objects across frames, new detections must be associated with existing map landmarks. 
The baseline DSP-SLAM uses a uniform distance threshold, which ignores semantic scale diversity and causes association errors (Fig.~\ref{fig:map_objects_association}(b-c-d)). 
To resolve this, we introduce a class-conditioned association, defining a unique matching radius $\mathrm{r}_c$ for each class $c$ based on its average ground-plane footprint diagonal: $\mathrm{r}_c = \sqrt{l_c^2 + w_c^2}$, where $l_c$ and $w_c$ are the length and width.
\par
The process iterates through currently tracked objects $\mathcal{M}$ and validated keyframe detections $\mathcal{D}_{\text{valid}}^{(k)}$ (Line 5). 
For each map object $m \in \mathcal{M}$ of class $c$, we identify the closest candidate detection $d^*_c$ of the same class (Line 6). 
Association is confirmed if their ground-plane distance, $\| \mathbf{p}_m - \mathbf{p}_{d^*_c} \|_2$, is less than $\mathrm{r}_c$ (Lines 8, 9).
\par
To handle reappearing objects, we perform a second global association against all map objects before creating a new instance. 
This prevents duplicates when objects are re-detected after tracking loss or during loop closures. 
Finally, any unassociated detection in $\mathcal{D}_{\text{valid}}^{(k)}$ is instantiated as a new landmark (Lines 12, 13, Fig.~\ref{fig:map_objects_association}(a)). 
This geometry-aware strategy improves robustness by reducing failed associations for large objects and preventing false matches in crowded scenes (Fig.~\ref{fig:map_objects_association}(c-d)).
\subsection{Class-Specific Shape Reconstruction}
\label{sec:multiclass_deepsdf}
Our shape reconstruction builds upon DeepSDF~\cite{park2019deepsdf}, a learned, continuous representation of a signed distance function (SDF) conditioned on a latent shape code $\mathbf{z}$. The baseline DSP-SLAM~\cite{wang2021dsp} uses, but not strictly bound to, DeepSDF to optimize an object's pose and shape from its observed point cloud and 2D mask. However, naively extending this by training a single, monolithic decoder for multiple object classes can lead to cross-class interference and overfitting. To mitigate this risk, we use a modular multi-decoder approach, maintaining a set of independent, pre-trained DeepSDF decoders, $\{G_c\}_{c \in \mathcal{C}}$, one for each supported object class.
\par
At runtime, when an object with semantic label $c$ is associated, the system dynamically selects the corresponding decoder $G_c$. The subsequent optimization of the object's pose, $T_{co}$, and shape code, $\mathbf{z}_c$, then proceeds as defined in the baseline (Lines 14, 15). 
By initializing and constraining this optimization to the correct class-specific manifold via $G_c$, the process converges faster and yields a more representative geometry.
This modular design makes our framework inherently extensible; new object classes can be supported by adding a pre-trained decoder without retraining the entire network. 
While synthetic priors introduce a synthetic-to-real gap, our network-agnostic design allows seamless integration of future domain-adapted models.
\section{Asynchronous Back-End for Real-Time Performance}
\label{sec:asynch_recons}  
% start writing here
%
%
\begin{figure}
\centerline{\includegraphics[width=3.4in] {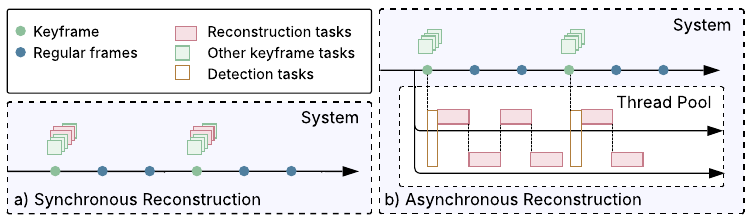}} % 3.4
\caption{(a) Synchronous reconstruction, (b) Asynchronous reconstruction.}
\label{fig:asynchronous_vs_synchronous}
\end{figure}
Prior implicit methods like DSP-SLAM are limited by keyframe processing bottlenecks, restricting the system to approximately 10--15~fps for stable performance. This latency stems from synchronously executing computationally intensive detection and reconstruction tasks within the main tracking and mapping threads. This concentrated load at each keyframe stalls all other SLAM operations (Fig.~\ref{fig:asynchronous_vs_synchronous}(a)). While manageable in single-class scenarios at low frame rates, this approach fails when handling multiple object classes, as multi-class reconstruction is inherently demanding. Faster detectors can reduce tracking latency, but faster shape decoders~\cite{sitzmann2020metasdf, chou2022gensdf} do not resolve the fundamental problem of concentrated load. Inspired by multi-threaded frameworks for real-time data processing~\cite{jana22020}, we introduce an Asynchronous Reconstruction (AR) pipeline that decouples object reconstruction from the time-critical mapping thread, distributing the workload across multiple frames (Fig.~\ref{fig:asynchronous_vs_synchronous}(b)).
\par
Our asynchronous workflow is summarized in Algorithm~\ref{alg:async_recon} and managed by a two-worker thread pool, a thread-safe task queue, and a results queue. The local mapping thread acts as the \textit{producer} (Line 2). When an object reconstruction is required, instead of waiting for the final mesh, it creates a placeholder initialized with the 3D bounding box detection. It then packages the necessary reconstruction data (\textit{e.g.}, point cloud, 2D mask, and current pose) into a task $\mathcal{T}$ and pushes it to the task queue (Lines 4--6), instantly freeing the thread. The worker threads act as \textit{consumers} (Line 13): available threads retrieve tasks, perform the computationally expensive DeepSDF optimization and mesh extraction, and place the result $\mathcal{R}$ (\textit{e.g.}, optimized latent code, pose, and 3D mesh) into the results queue (Lines 15--17). The local mapping thread periodically checks for available results (Line 8), dequeuing and integrating them into the global map (Lines 9, 10) to update the object mesh. Detection is prioritized over reconstruction; task processing pauses when new detections are required.
\begin{algorithm}
\caption{Asynchronous Reconstruction Pipeline}
\label{alg:async_recon}
\begin{algorithmic}[1]
    \State \textbf{shared data:} \textit{TaskQueue}, \textit{ResultQueue}
    \Statex
    \Procedure{LocalMapping}{keyframe}
        \State Process(keyframe)
        \If{reconstruction\_needed(keyframe)}
            \State $\mathcal{T} \gets \text{CreateTask(keyframe)}$
            \State \textit{TaskQueue}.push($\mathcal{T}$) \Comment{Add task to queue}
        \EndIf
        \Statex \Comment{Process any completed results without blocking}
        \While{\textbf{not} \textit{ResultQueue}.empty()}
            \State $\mathcal{R} \gets \textit{ResultQueue}.pop()$
            \State UpdateMap($\mathcal{R}$)
        \EndWhile
    \EndProcedure
    \Statex
    \Procedure{Worker}{} \Comment{Runs in a parallel thread pool}
        \While{True}
            \State $\mathcal{T} \gets \textit{TaskQueue}.pop()$ \Comment{Blocks until a task is available}
            \State $\mathcal{R} \gets \text{ReconstructShape}(\mathcal{T})$
            \State \textit{ResultQueue}.push($\mathcal{R}$)
        \EndWhile
    \EndProcedure
\end{algorithmic}
\end{algorithm}
\par
This approach not only ensures that the back-end mapping is never blocked by heavy reconstruction tasks, but also makes object insertion into the map instantaneous through these bounding box placeholders. Consequently, this preserves the real-time capability of the system and provides immediate spatial awareness for downstream planning, avoiding the risks associated with reconstruction delays even when modeling numerous complex objects.

% # # # # # # # # # # # # # # # # # # # # # # # # # # # # # # # # # # #

\section{Fisheye-LiDAR Depth Integration}
\label{sec:lidar_cam_sync} 
The front-end of our system processes raw data from a fisheye camera and a LiDAR sensor into a metric-scale stream suitable for the main SLAM pipeline. We build upon the ORB-SLAM3 backbone to leverage its mature support for wide-angle fisheye camera models. However, using a monocular fisheye camera introduces the inherent challenge of scale ambiguity. 
\begin{figure}[t]
    \centering
    \includegraphics[width=\linewidth]{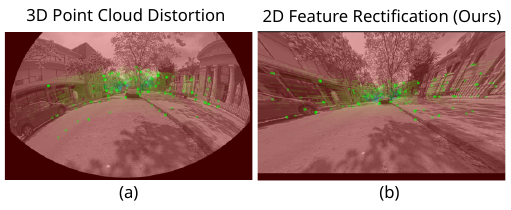}
    \caption{Fisheye-LiDAR depth association. (a) Distorting dense 3D LiDAR points into the raw fisheye space. (b) Our approach: rectifying sparse 2D visual features to a virtual pinhole space to minimize computational overhead. Transparent green dots represent LiDAR measurements, and solid squares represent SLAM features.  }
    \label{fig:projection_strategy}
\end{figure}
%-------------------- end figure-------------------
%------------------------------------------

To provide the monocular system with metric depth, we adapt the lightweight RGB-L method, which extracts depth from sparse LiDAR measurements. The primary challenge is the severe distortion of the fisheye lens, as the standard RGB-L projection logic assumes a pinhole camera model.

Instead of performing a computationally expensive full-image rectification, we leverage the native fisheye support in ORB-SLAM3 to associate LiDAR depth with the camera observations. This presents two possible projection strategies (illustrated in Fig.~\ref{fig:projection_strategy}): rectifying the extracted 2D visual features into an ideal pinhole space, or distorting the projected 3D LiDAR points into the raw fisheye image space.

We adopt the feature rectification approach due to its superior computational efficiency and its compatibility with the underlying ORB-SLAM3 codebase. Because the number of extracted visual features is significantly smaller than the dense set of LiDAR points falling within the camera's field of view, transforming this sparse feature set minimizes processing overhead compared to applying a complex distortion model to the entire point cloud. Specifically, for each feature point extracted from the distorted image, the system uses the Kannala-Brandt model to compute its corresponding undistorted coordinate. The LiDAR point cloud is then projected into this virtual pinhole space, allowing the system to accurately associate metric depth with each feature's rectified position.

\par
\section{Experiments \& Results} 
\label{sec:results}
\subsection{Experimental Setup} 
We evaluate our system on an in-house dataset\footnote{Dataset details will be provided upon paper acceptance.} collected by a vehicle navigating dense urban environments. This dataset is designed to be challenging, featuring scenarios with closely parked cars, motorcycles, and large buses that cause significant occlusion. Our data collection vehicle is equipped with four fisheye cameras, a 32-line LiDAR sensor, and a GNSS/INS unit to establish ground truth; however, our experiments utilize only the front-facing camera and the LiDAR stream.

To benchmark against prior work, we utilize the KITTI dataset. To demonstrate our system's versatility across different camera models, we also evaluate on the nuScenes \cite{caesar2020nuscenes} (pinhole) and CBNU \cite{Javed2022_CBNU} (fisheye) datasets. 
Our DSP-SLAM++ is compared against the original DSP-SLAM~\cite{wang2021dsp} only on the KITTI dataset, as the baseline lacks native fisheye camera support and cannot process our custom dataset. 
We specifically restrict our baseline comparison to DSP-SLAM, rather than primitive-based methods like CubeSLAM \cite{cubeslam}, to directly evaluate our architectural and multi-class improvements within high-fidelity implicit shape reconstruction.

Our system is evaluated on four object categories: cars, buses, motorcycles, and pedestrians. The front-end employs YOLO11x-seg and PointPillars as the 2D and 3D detectors, respectively. The reconstruction back-end utilizes a category-specific DeepSDF model for each class. 
Requiring only $\sim$7~MB of VRAM per model, this multi-decoder architecture is highly scalable. 
Furthermore, since objects are reconstructed sequentially, inference memory allocation remains independent of the total supported classes. 
These models are trained on 3D assets from ShapeNet~\cite{shapenet2015} and other online sources, which were preprocessed to ensure a consistent canonical orientation. All experiments were conducted on a machine with an Intel Core Ultra 9 275HX CPU and an NVIDIA GeForce RTX 5080 GPU.

We structure our evaluation as follows: first, we assess tracking accuracy and scale consistency against baseline methods. Next, we evaluate the quality of our multi-class 3D object reconstruction within static scenes. We then analyze the computational efficiency and real-time capabilities of our asynchronous architecture, concluding with ablation studies that isolate the impact of specific sensor modalities and system versatility.

% ----------------------------------------------
% ----------------------------------------------
% ----------------------------------------------
% ----------------------------------------------
\subsection{Tracking Accuracy and Scale Consistency}

\begin{figure}[t]
    \centering
    \includegraphics[width=\linewidth]{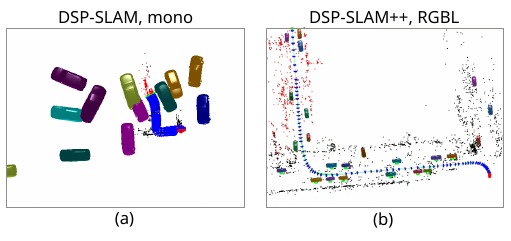}
    \caption{Impact of RGB-L integration on scale consistency. (a) Baseline DSP-SLAM (monocular) exhibits scale drift and erroneous object placement. (b) Our DSP-SLAM++ maintains metric scale for accurate trajectory and stable object mapping for monocular sensors.}
    \label{fig:scale_consistency}
\end{figure}

The baseline DSP-SLAM relies on stereo sensors for metric depth; strictly monocular setups introduce scale ambiguity, causing tracking drift and erroneous object placement (Fig.~\ref{fig:scale_consistency}). 
We evaluate our RGB-L module against the stereo configuration on the KITTI dataset, demonstrating our LiDAR-assisted approach maintains scale consistency.
As shown in Table~\ref{tab:tracking_accuracy_evaluation}, our system achieves comparable Absolute Trajectory Error (ATE), reported as mean and standard deviation across five independent runs. Furthermore, while the baseline fails on non-stereo datasets like nuScenes, our system leverages LiDAR depth inference to enable robust tracking, with full statistical results detailed in Table~\ref{tab:tracking_accuracy_evaluation}.

\begin{table}[t]
\centering
\caption{Tracking Accuracy: ATE Trans [m] / ATE Rot [$^\circ$]}
\label{tab:tracking_accuracy_evaluation}
% 1. Reduce padding between columns (default is 6pt)
\setlength{\tabcolsep}{3.5pt} 
% 2. Use a native small font instead of blindly scaling the whole box
\footnotesize 
\begin{tabular}{@{}lllccc@{}}
\toprule
\multirow{2}{*}{\textbf{Dataset}} & \multirow{2}{*}{\textbf{Seq.}} & \multirow{2}{*}{\textbf{Sensor}} & \textbf{DSP-SLAM} & \multicolumn{2}{c}{\textbf{DSP-SLAM++}} \\
\cmidrule(l){5-6}
 & & & (Baseline) & (w/o objects) & (w/ objects) \\ \midrule
\multirow{3}{*}{KITTI} 
 & 07 & Stereo & 0.83 / 0.94 & 0.81 / 0.92 & 0.86 / 1.02 \\
 & 07 & Mono   & X & X & X \\
 & 07 & RGB-L  & X & 0.85 / 0.96 & 0.90 / 1.01 \\ \midrule
\multirow{4}{*}{nuScenes} 
 & 127 & Mono & X & X & X \\
 & 127 & RGB-L & X & 0.85 / 0.84 & 0.93 / 0.77 \\
 & 152  & RGB-L & X & 1.92 / 1.10 & 1.66 / 1.78 \\
 & 916  & RGB-L & X & 1.44 / 4.70 & 1.49 / 4.87 \\ \midrule
Custom (Ours) 
 & 01  & RGB-L & X & 1.61 / 4.00 & 1.25 / 4.10\\ \bottomrule
\multicolumn{6}{l}{\scriptsize \textit{X: Indicates tracking failure, cluttered objects, or incompatibility.}}
\end{tabular}
\end{table}

%-------------------- begin figure-------------------
%-------------------- end figure-------------------

% \textbf{Objects as Landmarks.} Finally, we evaluate the benefit of incorporating object landmarks into the SLAM optimization. We compare the trajectory accuracy of our full system against a variant where objects are detected and reconstructed but not used as landmarks in the backend optimization. 
% Including multi-class objects as persistent landmarks improves the final ATE by XX\%, confirming their dual role in both creating a rich, semantic map and constraining the optimization to improve overall localization accuracy.

% ----------------------------------------------

\subsection{Multi-Class 3D Object Reconstruction}

\begin{figure}[t]
    \centering
    \includegraphics[width=\linewidth]{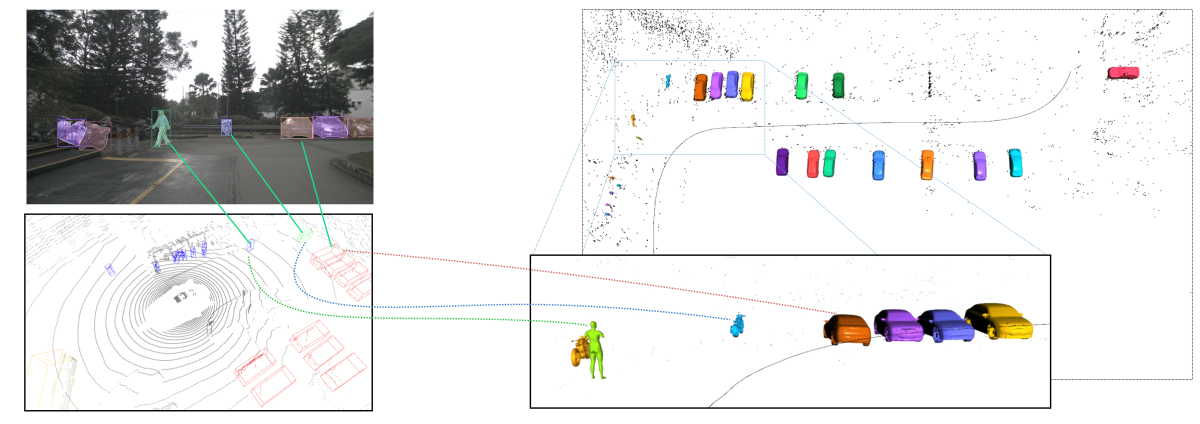}
    \caption{Generated multi-class 3D object map with overlaid trajectories and illustrated detections for nuScenes dataset (scene-0916).}
    \label{fig:multi_class_map}
\end{figure}

We qualitatively evaluate our framework on our in-house dataset and nuScenes to demonstrate its capacity for building rich, semantically aware maps. As illustrated in Fig.~\ref{fig:multi_class_map}, the system generates a large-scale map with the estimated trajectory overlaid, confirming that diverse object classes--cars, buses, motorcycles, and pedestrians--are accurately anchored in metric space. This multi-class representation provides a significantly richer world model for downstream tasks compared to single-class baselines.
\par
The system demonstrates robust shape reconstruction for voluminous, rigid bodies such as cars and buses. Furthermore, the class-aware association pipeline effectively handles cluttered environments, successfully disambiguating closely spaced instances, such as parked motorcycles (Fig.~\ref{fig:custom_dataset_map}). However, we acknowledge visual limitations when generating thin or articulated objects, specifically pedestrians. Because our system-level approach leverages existing shape priors, reconstruction fidelity for these challenging classes depends heavily on the observed LiDAR point cloud density and the pre-trained model quality. Consequently, for distant or heavily occluded instances where only sparse points are captured, the resulting meshes may lack fine detail or appear incomplete.

%-------------------- begin figure-------------------
%-------------------- end figure-------------------

\subsection{Computational Efficiency and Asynchronous Performance}

\begin{table*}[t]
\centering
\caption{Computational Efficiency (RGB-L Mode): Synchronous (SR) vs. Asynchronous (AR) Reconstruction. SC is for Single-Class, and MC is for Multi-Class Detection and Reconstruction.}
\label{tab:efficiency}
\renewcommand{\arraystretch}{1.2} 
\begin{tabular}{@{}llcccccc@{}}
\toprule
\multirow{2}{*}{\textbf{Dataset}} & \multirow{2}{*}{\textbf{Pipeline}} & \textbf{Detect. Latency} & \textbf{Object Latency} & \textbf{Max Obj. Latency} & \textbf{KF BA Latency} & \textbf{Map Objects} & \textbf{System FPS} \\
 & & \textbf{Mean $\pm$ Std [ms]} & \textbf{Mean $\pm$ Std [ms]} & \textbf{[ms]} & \textbf{Mean $\pm$ Std [frames]} & \textbf{[count]} & \textbf{[Hz]} \\ \midrule
\multirow{2}{*}{KITTI-SC} 
 & SR & \textbf{82 $\pm$ 6} & 106 $\pm$ 81 & 520 & 1.5 $\pm$ 0.8 & 96 & \textbf{27.3} \\
 & \textbf{AR (Ours)} & 86 $\pm$ 10 & \textbf{78 $\pm$ 47} & \textbf{483} & \textbf{1.3 $\pm$ 0.5} & \textbf{108} & 23.5 \\ \midrule
\multirow{2}{*}{Custom-SC} 
 & SR & 84 $\pm$ 11 & 57 $\pm$ 74 & 554 & 1.9 $\pm$ 1.3 & 16 & 25.5 \\
 & \textbf{AR (Ours)} & \textbf{81 $\pm$ 8} & \textbf{22 $\pm$ 23} & \textbf{146} & \textbf{1.2 $\pm$ 0.5} & \textbf{27} & \textbf{27.4} \\ \midrule
\multirow{2}{*}{Custom-MC} 
 & SR & \textbf{90 $\pm$ 11} & 82 $\pm$ 81 & 566 & 2.5 $\pm$ 1.5 & 43 & 25.1 \\
 & \textbf{AR (Ours)} & \textbf{90 $\pm$ 11} & \textbf{26 $\pm$ 25} & \textbf{147} & \textbf{1.2 $\pm$ 0.5} & \textbf{71} & \textbf{25.4} \\ \bottomrule
\end{tabular}
\end{table*}

We evaluate the computational efficiency of our Asynchronous Reconstruction (AR) pipeline against the baseline Synchronous Reconstruction (SR) approach. 
In SR, shape optimization is coupled with the local mapping thread, creating a severe bottleneck when processing multiple objects. We measure this delay using Keyframe Bundle Adjustment (BA) Latency, which quantifies the number of regular frames processed by the tracking thread while the mapping thread remains locked. This latency introduces a twofold problem: it forces the system to skip Local Bundle Adjustment (LBA) to catch up, and it forces the tracking thread to continue operating (\textit{e.g}., processing 5 frames) against raw, unrefined map points triangulated from initial poses. Tracking against these unoptimized points compounds trajectory errors. Our AR pipeline resolves this by decoupling the computationally expensive shape optimization, ensuring immediate LBA execution and robust tracking against a continuously refined map.

The comparative performance results across the KITTI and our custom datasets, averaged over five independent runs, are detailed in Table~\ref{tab:efficiency}. 
Under the SR paradigm in the custom Multi-Class (MC) scenario, the system exhibits massive mapping latency peaks (> 500 ms), causing the mapping thread to lag behind the tracking thread by a maximum of 15 frames. This severe desynchronization triggers frequent LBA skipping and forces prolonged tracking against unrefined map points. Conversely, our AR pipeline effectively distributes the computational load, cutting the mean map latency to one-third and reducing the maximum Keyframe BA latency to just 3 frames. By isolating the heavy object reconstruction backend (while maintaining a consistent detection latency of 90 ms), our architecture successfully minimizes tracking against raw points and prevents LBA skipping, enabling robust, real-time multi-class object SLAM.

% \textcolor{blue}{Pending: mention exporting detector weights to unblock the tracking thread.} 

% \begin{figure*}
%   \centering
%   \includegraphics[width=1.0\linewidth]{figures/multli_class_reconstruction.png}
%   \caption{
%     Qualitative shape and pose results, with reconstructed meshes that match the vehicles arrangement and orientations in the fisheye images:
%     (a) Multiple car objects, illustrating standard performance.
%     (b) Multiple cluttered motorcycles, illustrating the effectiveness of our class-consistent projection and disambiguation mechanisms.
%     (c)-(d) Objects of different classes in the same frame, illustrating multiple class reconstruction.
%     \textcolor{blue}{draft, include a row for single class case.}}
%   \label{fig:multli_class_reconstruction}
% \end{figure*}

\subsection{Sensor Modality and Versatility}

\begin{figure*}
\centering
\includegraphics[width=1.0\linewidth]{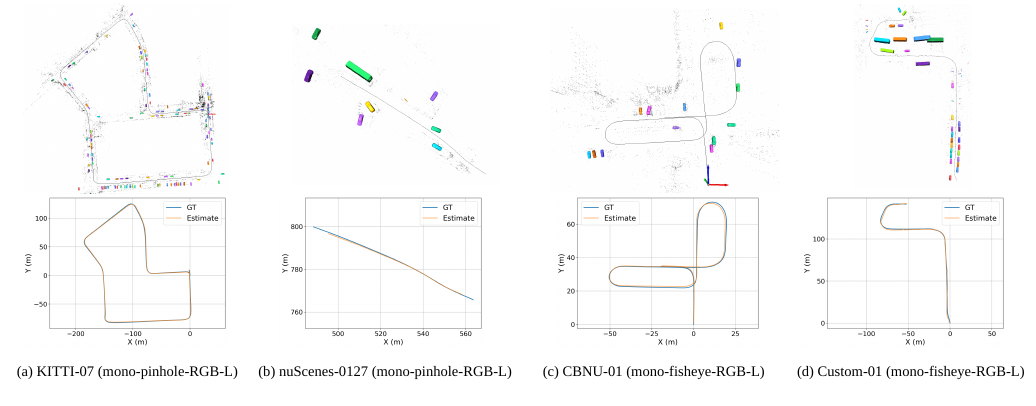}
\caption{Qualitative results across four diverse datasets. Top: 3D map reconstruction with object landmarks. Bottom: Estimated trajectories versus ground truth.}
\label{fig:system_validation}
\end{figure*}

To evaluate the versatility of our framework across sensor modalities, we demonstrate its performance on datasets featuring wide-angle fisheye cameras, specifically our custom and CBNU datasets. Map and trajectory tracking across these datasets are illustrated in Fig.~\ref{fig:system_validation}. While standard pinhole cameras suffer from a limited field-of-view in dense urban environments, fisheye sensors capture broader context at the cost of severe lens distortion. Our system seamlessly integrates the fisheye camera model within the tracking and mapping threads, ensuring objects are robustly detected and accurately anchored in the 3D map.

\begin{figure}[t]
    \centering
    \includegraphics[width=\linewidth]{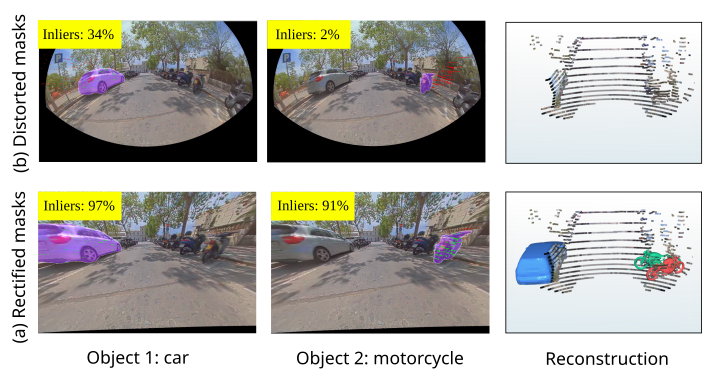}
    \caption{Mask rectification effect on confirmed detections and reconstruction.}
    \label{fig:mask_rectification_effect}
\end{figure}

A critical challenge in integrating LiDAR depth with fisheye imagery is accurately associating 3D point clouds with 2D instance masks. Direct projection using standard pinhole assumptions fails under high distortion, causing severe misalignment between physical geometry and semantic boundaries. To resolve this, our system performs explicit mask rectification to ensure correct projection of the point cloud onto the image plane. As shown in Fig.~\ref{fig:mask_rectification_effect}, this is crucial; without rectification, background depth points are erroneously assigned to foreground masks, causing corrupted shape initializations or missed associations. Successful deployment on the CBNU dataset validates that explicitly handling distortion allows the framework to generalize across diverse multi-sensor automotive setups.

Furthermore, mask rectification must be lightweight to maintain efficiency across different lens types and avoid introducing latency to the main tracking thread. Instead of rectifying the entire image grid, our approach extracts the contour of each predicted mask. We then apply the fisheye rectification model strictly to this dense set of perimeter points. Finally, the fully rectified mask is reconstructed by rendering the undistorted boundary as a filled polygon, maintaining precise shape consistency comparable to computationally expensive pixel-by-pixel rectification.
%-------------------- begin figure-------------------

%-------------------- end figure-------------------
%------------------------------------------

% -----------------------------------------------
% -----------------------------------------------
% Test 4: Multi-Class Detection and Reconstruction}
%-------------------- begin figure-------------------
%%% 3.34in is the maximum width you can have for a figure
\section{Conclusion} \label{sec:conclusion}

In this paper, we presented DSP-SLAM++, a framework that successfully addresses the trade-offs in object-aware SLAM by enabling real-time, multi-class, high-fidelity reconstruction on practical automotive sensor suites. 
Our system achieves this goal through three key contributions: multi-class detection, class-aware association, and a modular multi-decoder architecture for class-specific reconstruction, a novel asynchronous pipeline to resolve performance bottlenecks, and a fusion front-end for fisheye-LiDAR data.
Our experiments validate this approach, demonstrating a significant reduction in object reconstruction induced latency while maintaining high tracking accuracy compared to the state-of-the-art. This work makes high-fidelity object SLAM a more practical and accessible tool for real-world robotics applications like autonomous driving.

% While our system demonstrates strong performance, it has several limitations that open avenues for future research. The quality of the DeepSDF reconstruction is dependent on the density of the input point cloud; consequently, the system struggles to generate fine shapes for partially occluded or distant objects. Furthermore, the system's overall performance is coupled to the upstream object detectors. Promising directions for future work include fusing the SLAM-generated point cloud with LiDAR data to enhance the reconstruction of sparsely observed objects and exploring methods to relax the LiDAR dependency to broaden the system's applicability.

While our system performs robustly, specific limitations present clear avenues for future research. Reconstruction quality depends heavily on input point cloud density, resulting in degraded performance when handling sparse observations from occluded or distant objects. Additionally, system performance remains tightly coupled to the reliability of upstream detectors. Future work will address these issues by fusing SLAM-generated map points with LiDAR data to improve sparse reconstructions, integrating object tracking and motion forecasting to handle dynamic environments, and exploring methods to relax the LiDAR dependency to broaden the system's applicability.

\par
\section*{ACKNOWLEDGMENT}

This work is supported by the XXX project, (grant number XXX-XXXX).
% the DIDYMOS-XR Horizon Europe project (grant number 101092875–DIDYMOSXR,www.didymos-xr.eu).
%the University Research Board (URB) at the American University of Beirut (AUB).
%%%%%%%%%%%%%%%%%%%%%%%%%%%% bibliography %%%%%%%%%%%%%%%%%%%%%%%%%%%%
\bibliographystyle{ieeetr}
\bibliography{Ref/references}
\end{document}